\documentclass[letterpaper, 10 pt, conference]{ieeeconf}  % Comment this line out if you need a4paper

\IEEEoverridecommandlockouts                              % This command is only needed if 
\usepackage{graphics} % for pdf, bitmapped graphics files
\usepackage{epsfig} % for postscript graphics files
\usepackage{mathptmx} % assumes new font selection scheme installed
\usepackage{times} % assumes new font selection scheme installed
\usepackage{amsmath} % assumes amsmath package installed
\usepackage{amssymb}  % assumes amsmath package installed
\usepackage{subfigure}
\usepackage{todonotes}
\usepackage{multirow}

\usepackage{subfloat}

\usepackage{breqn}
\usepackage{graphicx}

\usepackage{hyperref}
\hypersetup{colorlinks=true,urlcolor=blue,citecolor=red}

\usepackage{caption}
\newcommand{\norm}[1]{\left\lVert#1\right\rVert} % Command for auto adjust norm!
\newcommand{\etal}{\textit{et al}. }

\mathchardef\mhyphen="2D % Define a "math hyphen"

\title{\LARGE \bf
Unsupervised Learning of Dense Optical Flow, Depth and Egomotion from Sparse Event Data 
}

\author{Chengxi Ye$^{1*}$, Anton Mitrokhin$^{1*}$, 
\\Cornelia Ferm\"{u}ller$^{1}$, James A. Yorke$^{2}$, Yiannis Aloimonos$^{1}$% <-this % stops a space
\thanks{$^{*}$ The authors contribute equally to this work.}
\thanks{$^{1}$University of Maryland Institute for
Advanced Computer Studies,         
College Park, MD 20742, USA.
        {\tt\small E-mails: cxy@umd.edu, amitrokh@umd.edu, fer@umiacs.umd.edu, yiannis@cs.umd.edu}}%
\thanks{$^{2}$Institute for Physical Science and Technology, University of Maryland,
        College Park, MD 20742, USA.
        {\tt\small E-mail:yorke@umd.edu}}%
}

\begin{document}

\maketitle
\thispagestyle{empty}
\pagestyle{empty}

%%%%%%%%%%%%%%%%%%%%%%%%%%%%%%%%%%%%%%%%%%%%%%%%%%%%%%%%%%%%%%%%%%%%%%%%%%%%%%%%
\begin{abstract}
% addresses the problem at multiple resolutions - PROBLEM OF WHAT?
% A feature decorrelation technique is introduced to improve the training of the network. - WE ADDITIONALLY INTRODUCE... 'improve the training' - WHAT DOES IT AMOUNT TO?

In this work we present a lightweight, unsupervised learning pipeline for \textit{dense} depth, optical flow and egomotion estimation from sparse event output of the Dynamic Vision Sensor (DVS). To tackle this low level vision task, we use a novel encoder-decoder neural network architecture - ECN.

%Our pipeline utilizes a novel multi-scale multi-level neural network architecture. A new feature decorrelation technique is introduced to train the network.
%To tackle this low level vision task, we use a novel encoder-decoder neural network architecture that incorporates multi-scale  features. Furthermore, a feature decorrelation technique is introduced to improve the training of the network. %A non-local sparse smoothness constraint is used to alleviate the challenge of data sparsity.

Our work is the first monocular pipeline that generates dense depth and optical flow from sparse event data only. The network works  in self-supervised mode and has  just 150k parameters. We evaluate our pipeline on the MVSEC self driving dataset and present  results for depth, optical flow and and egomotion estimation. Due to the lightweight design, the inference part of the network runs at 250 FPS on a single GPU, making the pipeline ready for realtime robotics applications. Our experiments demonstrate significant improvements upon previous works that used deep learning on event data, as well as the ability of our pipeline to perform well during both day and night.

\vspace{1.0\baselineskip}

\textit{Keywords -- event-based learning, neuromorphic sensors, DVS, autonomous driving, low-parameter neural networks, structure form motion, unsupervised learning}

\end{abstract}

\section*{Supplementary Material}

The supplementary video, code, trained models, appendix and a preprocessed dataset will be made available at \url{http://prg.cs.umd.edu/ECN.html}.

%%%%%%%%%%%%%%%%%%%%%%%%%%%%%%%%%%%%%%%%%%%%%%%%%%%%%%%%%%%%%%%%%%%%%%%%%%%%%%%%
\section{INTRODUCTION}
% Event-based processing in general:its advantages and challenges

%Visual motion is evolutionary the oldest and most important cue for encoding information about the 3D motion and spatial geometry of a scene. Even the most primitive animals, such as insects and reptiles, use visual motion  to interpret the space-time geometry surrounding them. Yet, even the most advanced Computer Vision algorithms are no match for the capabilities of biological systems.

With recent advances in the field of autonomous driving, autonomous platforms are no longer restricted to research laboratories and testing facilities - they are designed to operate in an open world, where reliability and safety are key factors. Modern self-driving cars are often fitted with a sophisticated sensor rig, featuring a number of LiDARs, cameras and radars, but even those undoubtedly expensive setups are prone to misperform in difficult conditions - snow, fog, rain or at night.

Recently, there has been much progress in imaging sensor technology, offering alternative solutions to scene perception.  A neuromorphic imaging device, called Dynamic Vision Sensor (DVS)~\cite{lichtsteiner_latency_2008}, inspired by the transient pathway of mammalian vision, can offer exciting alternatives for visual motion perception. The DVS does not record image frames, but instead - the changes of lighting occurring independently at every camera pixel. Each of these changes is transmitted asynchronously and is called an \textit{event}. %The design of this sensor is inspired by the human retina and this allows the sensor to accommodate for a 
By its design, this sensor accommodates a large dynamic range, provides high temporal resolution and low latency -- ideal properties for applications where high quality motion estimation and tolerance towards challenging lighting conditions are desirable.
%The price for these properties is indeed heavy -
%event-based sensors produce a lot of noise, its resolution is relatively low and its data - typically referred to as \textit{event cloud} - is
%asynchronous and sparse. 
The DVS  offers  new opportunities for robust visual perception so much needed in autonomous robotics, but challenges associated with the sensor output, such as high noise, relatively low spatial resolution and sparsity,  ask for  different visual processing approaches.

\begin{figure}[t]
\begin{center}
\includegraphics[width=0.95\columnwidth]{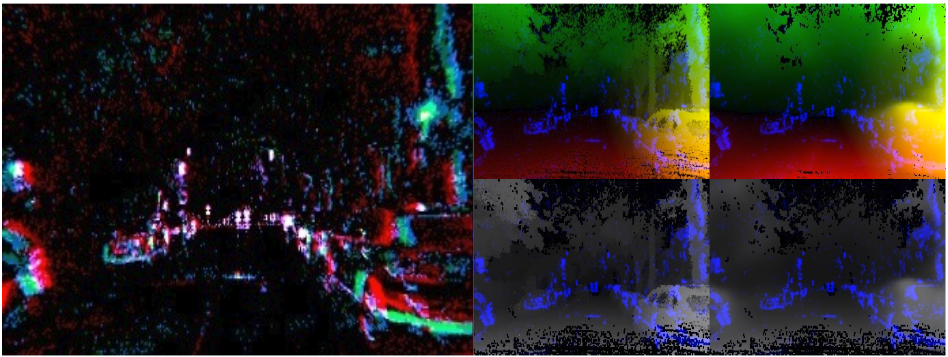}
\end{center}
\vspace{-0.5\baselineskip}
   \caption{\textit{\small{Optical flow and depth inference on sparse event data in night scene: event camera output (left), ground truth (middle column), network output (right) (top row - flow, bottom row - depth). The event data is overlaid on the ground truth and inference images in blue. Note, how our network is able to `fill in' the sparse regions and reconstruct the car on the right.}}}
   %Best viewed in color.
   \vspace{-1.0\baselineskip}
\label{fig:teaser}
\end{figure}
%\end{figure*}

In this work we introduce a novel lightweight encoding-decoding neural network architecture - the  Evenly-Cascaded convolutional Network (ECN) to address the problems of event data sparsity for depth, optical flow and egomotion estimation in a self driving car setting. Despite having just 150k parameters, our network is able to generalize well to
different types of sequences.
%on different types of sequences, as well as transfer from day  to night sequences. 
The simple nature of our pipeline allows it to run at more than 250 inferences per second on a single NVIDIA 1080 Ti GPU. We perform ablation studies using the SfMlearner architecture \cite{zhou2017unsupervised} as a baseline and evaluate different normalization techniques (including our novel \textit{feature decorrelation}) to show that our model is well suited for event data.

We demonstrate superior generalization  to  low-light scenes. Fig. \ref{fig:teaser} shows an example featuring night driving - the network trained on a day light scene was able to predict both depth and flow even with a low event rate and abundance of noise. This is facilitated by our event-image representation: instead of  the latest event timestamps, we use the average timestamp of the  events generated at a given pixel. The averaging helps to reduce the noise without losing the timestamp information. Moreover, we use multiple slices as  input to our model to better preserve the 3D structure of the event cloud and more robustly estimate egomotion. The main contributions of our work can be summarized as:

%Traditionally, low level tasks such as image segmentation, depth estimation and optical flow were solved by utilizing low level features at multiple resolutions. The recently introduced  deep neural networks have  an encoder-decoder architecture and they address these low level tasks with high-level features universally used in deep networks~\cite{DBLP:conf/miccai/RonnebergerFB15,zhou2017unsupervised,DFIB15}.They do not use low level features at multiple resolutions. In this work we introduce a novel encoding-decoding neural network architecture that utilizes both low-level and high-level features and addresses the final task from coarse-to-fine using multiple resolutions. We also utilize a sparse smoothness constraint, which is tailored for sparse data. 

\begin{itemize}
\item The first unsupervised learning-based approach to  structure from motion recovery using monocular DVS input.
\item Demonstrating  that dense, meaningful scene and motion information can be reliably recovered from sparse event data.
\item A new lightweight high-performance 
%encoder-decoder 
network architecture -- $ECN$.
\item A new normalization technique, called \textit{feature decorrelation}, which significantly improves training time and inference quality.
%\item A data representation (average time image), that improves robustness in difficult lighting conditions. 
\item Quantitative evaluation on the MVSEC dataset \cite{MVSEC} of dense and sparse depth, optical flow and egomotion.
\item A pre-processesed MVSEC \cite{MVSEC} dataset facilitating  further research on event-based SfM.
%\item Ablation study?
\end{itemize}

\section{Related Work}

\begin{figure}[t]
\begin{center}
\includegraphics[width=1.0\columnwidth]{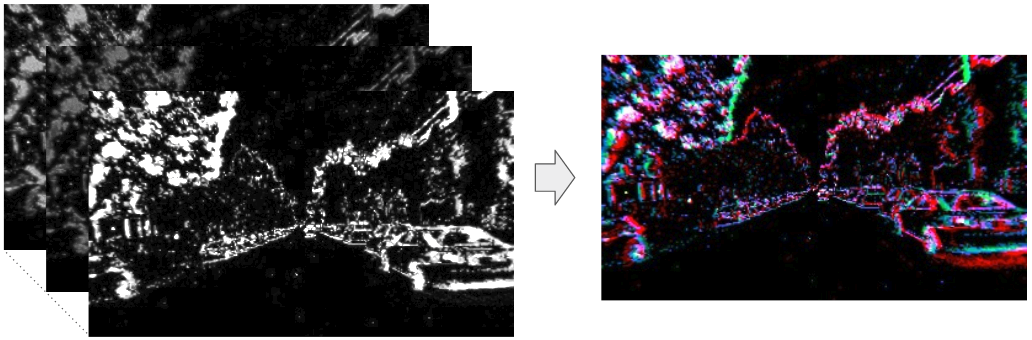}
\end{center}
\vspace{-0.5\baselineskip}
   \caption{\textit{A three-channel DVS data representation. The first channel represents the time image described in \cite{iROSBetterFlow}. The second and third channels represent the per-pixel positive and negative event counts. Best viewed in color.}}
   \vspace{-1.0\baselineskip}
\label{fig:representation}
\end{figure}

% To our knowledge, no previous work has presented an unsupervised deep learning framework for joint estimation of depth, and optical flow using a monocular event camera. All the related work either solves the joint estimation problem using classical approach or uses a learning approach tailored to one specific task. 

\subsection{Event-based Depth Estimation}
%  Over the years, several researchers have considered the problem of event-based monocular depth, motion and optical flow estimation either jointly or separately. \\
 
The majority of event-based depth estimation methods 
%\cite {6112233,kogler2011event,zhu2018realtime,
\cite{zhou2018semi,UpennSlices} use two or more event cameras. As our proposed approach uses only one event camera, we focus our discussion on monocular depth estimation methods. The first works on event-based monocular depth estimation were presented in \cite{BMVC2016_63} and \cite{10.1007/978-3-319-46466-4_21}. Rebecq \etal \cite{BMVC2016_63} used a space-sweep voting mechanism and maximization strategy to estimate semi-dense depth maps where the  trajectory is known. Kim \etal \cite{10.1007/978-3-319-46466-4_21} used probabilistic filters to jointly estimate the motion of the event camera, a 3D map of the scene, and the intensity image. More recently, Gallego \etal \cite{Gallego2018AUC} proposed a unified framework for joint estimation of depth, motion and optical flow. So far there has been no deep learning framework to predict depths from a monocular event camera.
 
\subsection{Event-based Optical Flow}
Previous approaches to image motion estimation used local information in event-space. The different methods adapt in smart ways one of the three principles known from frame-based vision, namely correlation \cite{delbruck_frame_2008,liu2017block}, gradient \cite{benosman_asynchronous_2012} and  local frequency estimation \cite{tschechne2014event}
%barranco2015bio}. 
The most popular approaches are gradient based - namely, to fit local planes to event clouds \cite{benosman_event_2014}
%,mueggler2015lifetime}. 
As discussed in \cite{barranco2014contour}, local event information is inherently ambiguous. To resolve the ambiguity Barranco et al. \cite{barranco2014contour} proposed to collect events over a longer time intervals and compute the motion from the trace of events at contours.
%that contours create when moving over multiple pixels. 

Recently, neural network approaches have shown promising results in various estimation problems without explicit feature engineering. Orchard and Etienne-Cummings \cite{6891162} used a spiking neural network
to estimate flow. Most recently, Zhu \etal \cite{zhu2018ev} released the MVSEC dataset \cite{MVSEC} and proposed self-supervised learning algorithm to estimate optical flow. Unlike \cite{zhu2018ev}, which uses grayscale information as a supervision signal, our proposed framework uses only events and thus can work in challenging lighting conditions.
%to predict optical flow.

% \subsection{Event-based Learning}

%     - Supervised - Maqueda et. al. 2018. Nguyen et al. 2017\\
%     - Self supervised - Zhu et al. 2017, Shedligeri et al. 2018 

\subsection{Self-supervised Structure from Motion}
The unsupervised learning framework for 3D scene understanding has recently gained popularity in frame-based vision research. Zhou et. al~\cite{zhou2017unsupervised} pioneered this line of work. The authors followed a traditional geometric modeling approach and built two neural networks, one for learning pose from single image frames, and one for pose from consecutive frames, which were self-supervised by aligning the frames via the flow. Follow-up works~
(e.g. \cite{ DBLP:journals/corr/abs-1802-05522}
%,DBLP:journals/corr/abs-1803-02276} 
have used similar formulations with better loss functions and networks, and recently
\cite{UpennSlices} proposed  SfM learning from stereo DVS data.
%After original archive publication of this paper, \cite{UpennSlices} proposed  SfM learning from stereo DVS data.  Our ECN network has better performance for 3D motion using only a single camera, and it generalizes from day to night scenes. 

\section{Methods}

\subsection{Ego-motion Model}
We assume that the camera is moving with  rigid motion with translational velocity $v = (v_x,v_y,v_z)^T$ and rotational velocity $\omega = (\omega_x, \omega_y,\omega_z)$, and that the camera intrinsic parameters  are provided.  Let $\mathbf{X}=(X,Y,Z)^T$ be the world coordinates of a point, and $\mathbf{x}= (x,y)^T$ be the corresponding pixel coordinates in the calibrated camera. Under the assumption of rigid motion, the image velocity $\mathbf{u} = (u,v)^T$ at $(x,y)^T$  is given as:
\begin{equation}
\resizebox{.9\hsize}{!}{$
\begin{pmatrix}
u\\v
\end{pmatrix}\\
=\frac{1}{Z}
\begin{pmatrix}
-1 &0 &x\\
0 &-1 &y
\end{pmatrix}
\begin{pmatrix}
v_x\\
v_y\\
v_z
\end{pmatrix}+
\begin{pmatrix}
xy &-1-x^2 &y\\
1+y^2 &-xy &-x
\end{pmatrix}
\begin{pmatrix}
\omega_x\\
\omega_y\\
\omega_z
\end{pmatrix}\\
= A \mathbf{p}
$}
\label{uv2}
\end{equation}
In words, given the inverse depth and the ego-motion velocities $\mathbf{v},\omega$, we can calculate the optical flow or pixel velocity at a point using a simple matrix multiplication (Equation~\ref{uv2}) Here $\mathbf{p}$ is used to denote the pose vector $(\mathbf{v},\omega)^T$, and $A$ is a $2\times6$ matrix. Due to scaling ambiguity in this formulation, depth $Z$ and  translation $(v_x, v_y, v_z)$ are computed up to a scaling factor. 
\subsection {Input Data Representation}
The raw data from the DVS sensor is a stream of events, which we treat as data of 3 dimensions. Each event encodes the  pixel coordinate $(x, y)$ and the  \textit{timestamp} $t$.
%, which convey the bulk of the information about the scene motion. 
In addition, it also carries information about its \textit{polarity} - a binary value that  disambiguates events generated on rising light intensity (positive polarity) and events generated on falling light intensity (negative polarity).

The 3D $(x, y, t)$ event cloud (within a small time slice), called event slice, is projected onto a plane and converted to a 3-channel image. An example of such image can be seen in Fig. \ref{fig:representation}. The two channels of the image are the per-pixel counts of positive and negative events. The third channel is the \textit{time image} as described in \cite{iROSBetterFlow} - each pixel consists of the average timestamp of the events generated on this pixel,
because the averaging of timestamps provides better noise tolerance.
%We argue that the averaging of the timestamps allows better noise  tolerance, allowing our pipeline to work in low-light conditions and handle cases of fast motion, where more recent events overwrite the previous ones. 
The neural network input consists of up to 5 such consecutive slice images to better preserve the timestamp information of the event cloud. 
\subsection {The Pipeline}

\begin{figure}[t!]
\begin{center}
\includegraphics[width=.95\columnwidth]{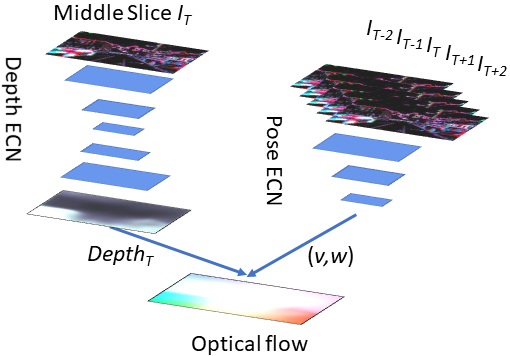}
\end{center}
\vspace{-0.5\baselineskip}
   \caption{\textit{\small{The depth network (left) with an encoder-decoder architecture is used to estimate scene depth. The  pose network (right) takes consecutive frames to estimate the translational velocity and rotational velocity with respect to the middle frame. Given the poses of neighboring frames and the depth of the middle frame, we calculate the optical flow. The neighboring frames are inversely warped to the middle frame and the warping difference provides the supervision loss.}}}
   \vspace{-1.0\baselineskip}
\label{fig:pipeline}
\end{figure}
The global network structure  is similar to the one proposed in \cite{zhou2017unsupervised}. It consists of a depth prediction component and a parallel pose prediction component, which both feed into a optical flow component to warp successive event slices. The loss from the warping error is backpropagated to train flow, inverse depth, and pose.
%We also use an explainability mask for locating regions that don't fit the rigid motion model. 

Our network components, instead of the   standard CNNs, are based on  our ECN network structure.
An (ECN based) encoding-decoding architecture  is used to estimate  scaled inverse depth $\frac{1}{Z}$ from a single slice of events.
%, rather than from normal RGB images~\cite{zhou2017unsupervised}.
To address the data sparsity,  we use bilinear interpolation, which propagates local information and fills in the gaps between events. A second  network, which  takes consecutive slices of signals, is used to derive $v$ and $\omega$.
%the translational velocity $v$ and rotational velocity $\omega$. % Under the rigid ego-motion assumption, the velocity of each pixel can be predicted from $v,\omega$ and $\frac{1}{Z}$ using simple matrix multiplication (Eq.~\ref{uv2}). 
Then, using the rigid motion and  inverse depth to predict the optical flow, neighboring slices at neighboring time instances $T+1, T+2$ and $T-1, T-2$ are warped to the  slice at $T$ (Fig.~\ref{fig:pipeline}). The $l^1$ loss is used to measure the difference between the warped events and the middle slice as the supervision signal.

\subsection {Evenly Cascaded Network Architecture}
\begin{figure}
\centering
\includegraphics[width=1.\columnwidth]{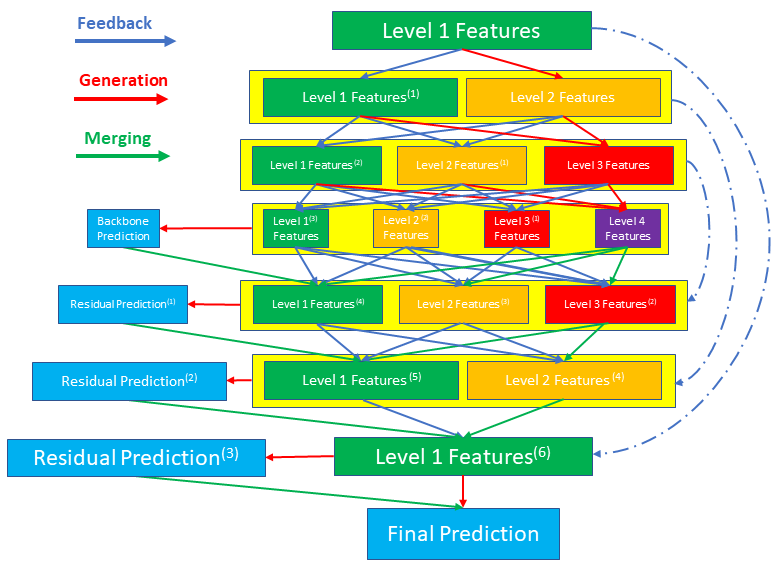}
\caption{\textit{\small{The encoder-decoder structure. Each encoding layer has two streams of feature maps. The feedback signal is added to the downsampled current features themselves to improve them (with improving times braced in the superscript). A set of high-level features are generated and concatenated with existing feature channels. The decoding stage performs reversely as the encoding stage, in which the feature maps expand in size. The highest-level features and the corresponding encoding layer (via skip link) are merged back into the lower-level features as modulation. We make predictions of depth at multiple scales using different features. A coarse, backbone depth is predicted then upscaled and refined using modulated lower-level features.}}}
\vspace{-1.0\baselineskip}
\label{fig:ecn}
\end{figure}

We use an encoder network %(Fig.~\ref{fig:ecn_encoder}) 
to estimate  pose from consecutive frames and an encoder-decoder network~\cite{DBLP:conf/miccai/RonnebergerFB15} (Fig.~\ref{fig:ecn}) to estimate the scaled depth. Next, we describe its main novelties.
%different novel computational steps, which we introduced to address specific challenges in the standard NN design.

\iffalse
\begin{figure}
\centering
\includegraphics[width=.8\columnwidth]{img/ecn_encoder.png}
\caption{\textit{\small{The evenly-cascaded encoder structure. Each encoding layer has two streams of feature maps. The feedback signal is added to the current features themselves to improve them (the improving times is braced in the superscript). A set of high-level features are generated and concatenated with existing feature channels.  We make pose predictions using all levels of features.}}}
\vspace{-1.0\baselineskip}
\label{fig:ecn_encoder}
\end{figure}
\fi

First, our transform of features takes biological inspiration from multi-stage information distillation, and incorporates feedback. %Our practical implementation resembles the celebrated cascade algorithm from  wavelet packet transforms~\cite{WaveletPackets}, where at each level of the transform signals are decomposed into two streams of a low and high frequency coefficients. 
Our encoding layers~\cite{DBLP:conf/bigdataconf/YeDMFA18} split the (layer) input  into two streams of features (Fig.~\ref{fig:ecn}): one incorporates `feedback' via residual learning~\cite{DBLP:journals/corr/HeZRS15} (convolution outputs are added to \textit{downsampled} existing features as modulation signals); the other directly generates a set of higher level features (convolution outputs are directly utilized). At the end of the encoding stage, the network has a multi-scale feature representation. This representation is used in our pose prediction.
%(see Fig.~\ref{fig:ecn_encoder})
%Our decoding stage is similar to the `merging' operation in wavelet reconstruction. 
In each decoding layer, the highest level features, together with the corresponding features in the encoding layers are convolved and added back to the \textit{upsampled} lower level features as modulation. At the end of the decoding stage, the network acquires a set of modulated high resolution low-level features. %It is important to point out that all modulation signals are added to the lower-level features as is common in residual learning. 

Our evenly-cascaded (EC) structure facilitates training by providing easily-trainable shortcuts in the architecture. The mapping $f$ from network input to output is decomposed into a series of progressively deeper, and therefore harder to train functions: $f=f_1+f_2+...+f_N$. The leftmost pathway (the green blocks) in Fig.~\ref{fig:ecn} contains the easiest to train, lowest-level features. A backbone pathway remains \textit{\textbf{unblocked}} by only going through downsampling and upsampling, promises the final gradient can be propagated to the first network layer. This construction alleviates the vanishing gradient problem in neural network training. More difficult-to-learn, modulation signals are added to this backbone pathway, allowing the network to selectively enhance and utilize multiple levels of features.

Second, 
%standard neural network downsampling and upscaling techniques, such as pooling and transposed convolutions~\cite{DBLP:conf/miccai/RonnebergerFB15}, are limited by integer scaling factors. Networks need to be carefully handcrafted according to the problem size. Furthermore, upscaling with transposed convolutions are  known to introduce unwanted `checkerboard artifacts.' To 
to tackle the challenges raised by sparse event data and evenly resize the features, we use bilinear interpolation. In the encoding layers, our network evenly downscales the  feature maps by a scaling factor of $(s < 1)$ to get coarser and coarser features until the feature size falls below a predefined threshold. In the decoding layers, the feature maps are reversely upscaled  by a scaling factor of $1/s$. The network construction is automatic and is controlled by the scaling factor. Bilinear interpolation propagates the sparse data spatially, facilitating  dense prediction of depth and optical flow.

\begin{figure}[t!]
\begin{center}
\includegraphics[width=0.475\textwidth]{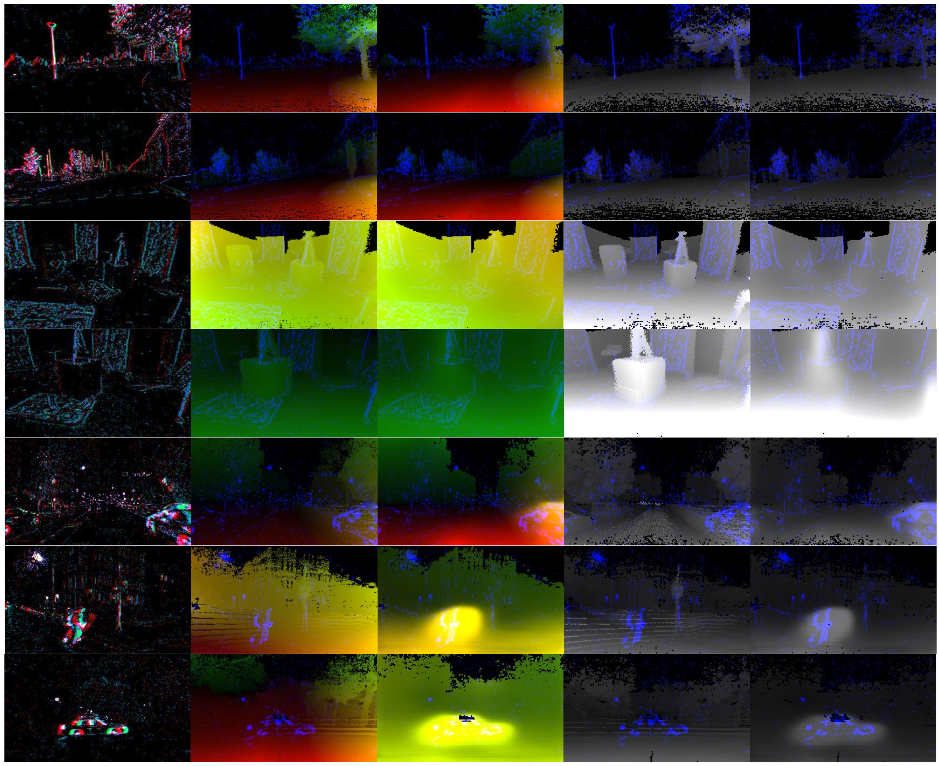}
\end{center}
\vspace{-0.5\baselineskip}
   \caption{\textit{\small{Qualitative results from our evaluation. The table entries from left to right: DVS input, ground truth optical flow, network output for flow, ground truth for depth, network output for depth. The event counts are overlaid in blue for better visualization. Examples were collected from sequences of the MVSEC \cite{MVSEC} dataset: (top to bottom) outdoor day 1, outdoor day 1, indoor flying 1, indoor flying 2, outdoor night 1, outdoor night 2, outdoor night 3. It can be seen that on the `night' sequences the ground truth is occasionally missing due to Lidar limitations but the pipeline performs well. Best viewed in color.}}}
   \vspace{-1.0\baselineskip}
\label{fig:big_pic}
\end{figure}

\subsection {Depth Predictions}
\label{sec:depth_prediction}

In the decoding stage, we make predictions from features at different resolutions and levels (Fig.~\ref{fig:ecn}). Initially, both high and low-level coarse features are used to predict a backbone depth map. The depth map is then upsampled with bilinear interpolation for refinement. Then the enhanced lower level features are used to estimate the prediction residue, which are usually also low-level structures. The residue is added to the backbone estimation to refine it. The final prediction map is therefore obtained through successive upsamplings and refinements. We also apply a smoothness constraint on the depth prediction.

Our pipeline is based on monocular vision and predicts depth up to a scale.
%The ground truth depth can recovered with binocular vision, as in a follow-up work~\cite{zhu2018ev}. 
In real world driving scenes, the mean depth value stays relatively stable. Taking advantage of this observation, we use  batch normalization before making the depth prediction, and this way  the predicted depths have similar range. Optionally, we also apply depth normalization~\cite{DBLP:journals/corr/abs-1712-00175} to strictly enforce that the mean depth remains a constant. 

\subsection {Feature Decorrelation}
Gradient descent training of neural networks can be challenging if the features are not properly conditioned, and features are correlated.
%For neural networks, the feature channels are usually correlated due to the interplay of channels. The signal amplitude can also be different due to layers of transforms.
Researchers have proposed normalization strategies~\cite{DBLP:journals/corr/IoffeS15,DBLP:journals/corr/abs-1803-08494} to account for the scale inconsistency problem. We proceed one step further with a decorrelation algorithm to combat the feature collinearity problem. 
%Straightforward decorrelation can be achieved by applying the inverse square root of the covariance matrix to the mean subtracted features. However, for neural network training, calculating inverse square roots of matrices at each iteration not only is computationally expensive but introduces instability.
We apply Denman-Beavers iterations~\cite{Denman:1976:MSF:2612816.2613024} to decorrelate the feature channels in a simple and forward fashion. Given a symmetric positive definite covariance matrix $C$, Denman-Beavers iterations start with initial values $Y_0=C$, $Z_0=I$. Then we iteratively compute: $Y_{k+1}=\frac{1}{2}Y_{k}(3I-Z_{k}Y_{k}),Z_{k+1}=\frac{1}{2}(3I-Z_{k}Y_{k})Z_{k}$. We then have $Z_{k}\xrightarrow{} C^{-\frac{1}{2}}$~\cite{DBLP:journals/corr/LinM17}. In our implementation, we evenly divide the features into 16 groups as proposed in group normalization~\cite{DBLP:journals/corr/abs-1803-08494}, and reduce the correlation between the groups by performing a few (1-10) Denman-Beavers iterations. We notice that a few iterations lead to significantly faster convergence and better results. 
%\subsection {Non-local Smoothness Penalty}
%To combat the sparsity in data, we utilize a sparsity constraint that promotes non-local information propagation: $Loss_{smooth}(I)=\sum_i \sum_{j\in N(i)} |I_j-I_i|^p =\sum_i \sum_{j\in N(i)} |I_j-I_i|^{p-2}|I_j-I_i|^2=\sum_i \sum_{j\in N(i)}  w_{ij}|I_j-I_i|^2$. Here the loss is applied on the first-order derivatives of the depth estimation, and we use a sparse penalty where $0<p= 1$. The complexity of the loss is quadratic in the neighborhood size. Acceleration techniques have been applied to reduce it to $O(1)$~\cite{DBLP:journals/corr/abs-1305-3971}.

%TODO: fix this table!
\renewcommand{\arraystretch}{1.3}
\begin{table*}[t]
\caption{\small{Evaluation of the optical flow pipeline}}
\vspace{-0.5\baselineskip}
\begin{center}
\resizebox{1.0\textwidth}{!}{\begin{tabular}{@{\extracolsep{0pt}}lcccccccccc}
\hline
 %& ''outdoor driving day'' & ''outdoor driving night'' & ''indoor flying 1'' & ''indoor flying 2'' & ''indoor flying 3'' \\
 & \multicolumn{2}{l}{outdoor day 1} & \multicolumn{2}{l}{outdoor night 1} & \multicolumn{2}{l}{outdoor night 2} & \multicolumn{2}{l}{outdoor night 3} & \multicolumn{2}{l}{indoor flying} \\
 \cline{2-3}\cline{4-5}\cline{6-7}\cline{8-9}\cline{10-11}
 & AEE & \% Outlier & AEE & \% Outlier  & AEE & \% Outlier  & AEE & \% Outlier  & AEE & \% Outlier  \\

\hline\hline
\textbf{$ECN$} & 0.35 & 0.04 & \textbf{0.49} & \textbf{0.82} & \textbf{0.43} & \textbf{0.79} & \textbf{0.48} & \textbf{0.80} & 0.21 & \textbf{0.01} \\
%\textbf{$EC{\mhyphen}SfM^2$} & \textbf{92.78\%} & \textbf{87.32\%} & \textbf{84.52\%} & \textbf{89.21\%} & \textbf{90.83\%} & \textbf{92.78\%} & \textbf{87.32\%} & \textbf{84.52\%} & \textbf{89.21\%} & \textbf{90.83\%} \\
\textbf{$ECN_{masked}$} & \textbf{0.30} & 0.02 & 0.53 & 1.1 & 0.49 & 0.98 & 0.49 & 1.1 & \textbf{0.20} & \textbf{0.01} \\ 
%\textbf{$EC{\mhyphen}SfM_{masked}^2$} & \textbf{92.78\%} & \textbf{87.32\%} & \textbf{84.52\%} & \textbf{89.21\%} & \textbf{90.83\%} & \textbf{92.78\%} & \textbf{87.32\%} & \textbf{84.52\%} & \textbf{89.21\%} & \textbf{90.83\%} \\
\textbf{$ECN_{erate}$} & 0.28 & 0.02 & 0.46 & 0.67 & 0.40 & 0.53 & 0.43 & 0.67 & 0.20 & 0.01 \\
\textbf{$Zhu18$ \cite{UpennSlices}} & 0.32 & \textbf{0.0} & - & - & - & - & - & - & 0.84 & 2.50 \\
\textbf{$EV{\mhyphen}FlowNet_{best}$ \cite{zhu2018ev}} & 0.49 & 0.20 & - & - & - & - & - & - & 1.45 & 10.26 \\
\textbf{$SfMlearner$} & 0.58 & 0.89 & 0.59 & 1.01 & 0.78 & 1.32 & 0.59 & 1.38 & 0.55 & 0.29 \\
\hline
\end{tabular}}
\vspace{-2.0\baselineskip}
\end{center}
\label{table:main_table}
\end{table*}
\renewcommand{\arraystretch}{1}

\section{Experimental Evaluation}
%The main contribution of our self-supervised learning framework lies in its ability to 
Our our self-supervised learning framework can infer both dense optical flow and depth from sparse event data. We evaluate our work on the MVSEC \cite{MVSEC} event camera dataset which, given a ground truth frequency of 20 Hz, contains over 40000 ground truth images.

The MVSEC dataset, inspired by KITTI \cite{KITTI1},
%, KITTI2}, 
 features 5 sequences of a car on the street (2 during the day and 3 during the night), as well as 4 short indoor sequences shot from a flying quadrotor. MVSEC was shot in a variety of lighting conditions and features low-light and high dynamic range frames which are often challenging for an analysis with classical cameras.

\begin{figure}[b]
\begin{center}
\includegraphics[width=1.0\columnwidth]{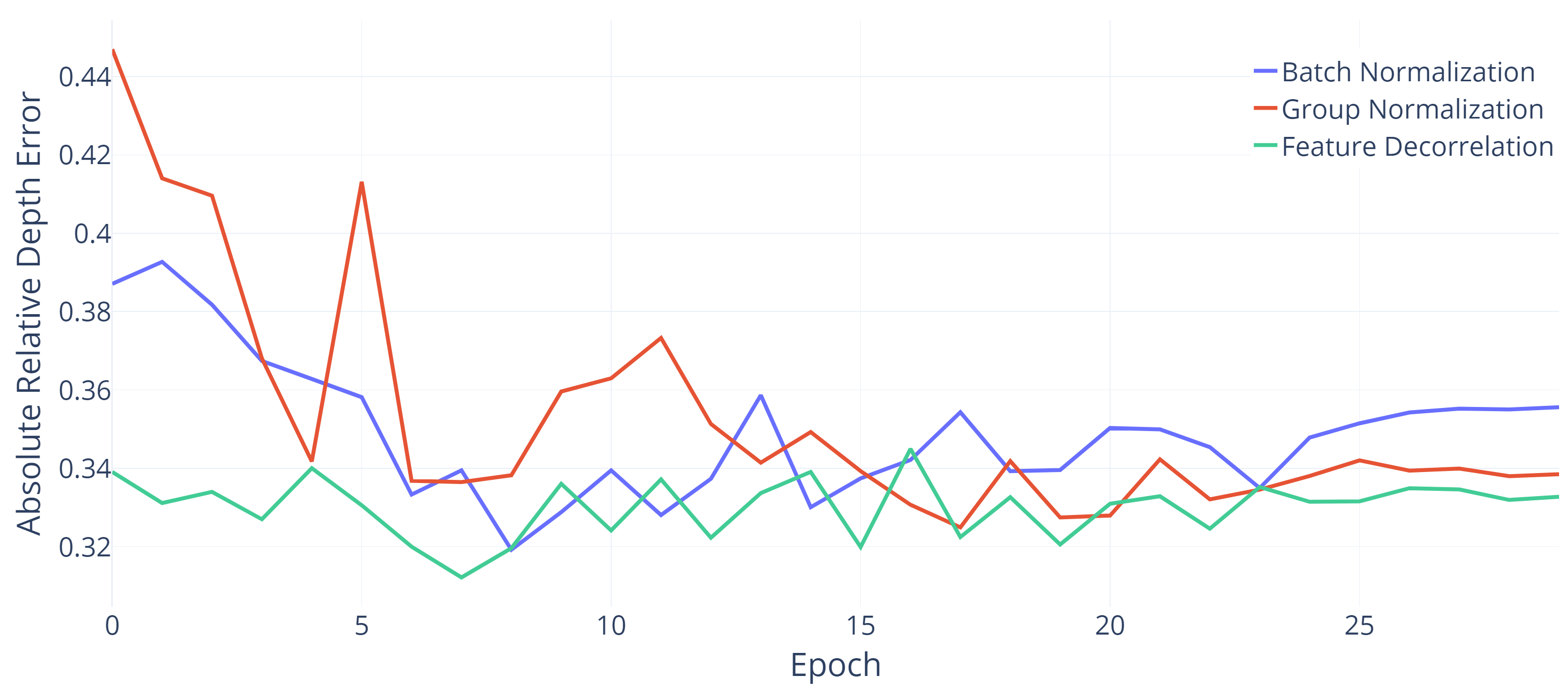}
\end{center}
\vspace{-1.0\baselineskip}
   \caption{\textit{\small{Comparison of Abs Rel Errors using different normalization methods on outdoor day 1 sequence (less is better).}}}
   \vspace{-0.5\baselineskip}
\label{fig:normalizations}
\end{figure}

\subsection{Implementation Details}
Our standard network architecture has scaling rates of $0.5$ and $2.0$ respectively for the encoding and decoding layers, and results in $5$ encoding/decoding layers.  Our depth network has $8$ initial hidden channels and expands with a growth rate of $8$. We halve these settings to $4$ for our pose network. The pose network takes $5$ consecutive event slices and predicts $6d$ pose vectors.
%of the neighboring slices with respect to the middle slice.  
We use $3\times 3$ convolutions, and the combined network has $150k$ parameters. We train the network with a batch size of $32$ and use the Adam optimizer with a learning rate of $0.01$. Interestingly, we notice that compared to the standard architecture of the SfMlearner, the learning rate is higher. Thus, the new design allows us to learn at a faster rate. The learning rate is annealed using cosine scheduling, and we let the training run for $30$ epochs. Our training takes $7$-minutes for each epoch using a single Nvidia GTX 1080Ti GPU. Our model using batch normalization runs at 250 FPS at inference as it has been heavily GPU optimized. The model using feature decorrelation runs at 40 FPS. The slow down is mainly due to  matrix multiplications in our customized layer which are less optimized for the GPU.

In the  experiments with the outdoor sequences, we trained the network using only the \textit{outdoor day 2} sequence with the hood of the car cropped. Our experiments demonstrate that our training generalizes well to the notably different \textit{outdoor day 1} sequence, as well as to the \textit{night} sequences. For the \textit{indoor} sequences, since they were too short to create a representative training set, we used $80\%$ of each sequence for training and evaluated on the remaining $20\%$. We would like to note that the \textit{outdoor night} sequences have occasional errors in the ground truth (see for example Fig. \ref{fig:big_pic} last three rows, or Fig. \ref{fig:failure_1}). All incorrect frames had to be manually removed for the evaluation. In our experiments, we use  fixed-width time slices of $1/40$-th of a second.

\subsection{Ablation Studies}
\subsubsection{Testing on the \textit{SfMlearner}}
%We provide the baseline results by training and evaluating the
As baseline we use the state-of-the-art \textit{SfMlearner} \cite{zhou2017unsupervised} on our data (event images). \textit{SfMlearner} has a fixed structure of 7 encoding and 7 decoding layers. It has 32 initial hidden channels and expands to 512 channels. Overall the model contains 33M parameters. SfMlearner is trained using Adam optimizer with a learning rate of $2e-4$ and a batch size of $4$. We replace the inputs with our event slices, and we include the evaluation results for flow and egomotion in tables \ref{table:main_table} and \ref{table:main_table_egomotion}.

\subsubsection{Normalization Methods}
We compare two normalization methods and our decorrelation method on the validation set portion of the outdoor day 2 sequence. We apply $5$ Denman-Beavers iterations in the decorrelation procedure. Compared with normalization methods, decorrelation leads to more thorough data whitening, and we have noticed this layer-wise whitening lead to faster convergence and lower evaluation loss (Fig.~\ref{fig:normalizations}).

\subsubsection{Visualizing a Shallow and Tiny Network}
\begin{figure}[h]
\centering
%\subfloat(a)
\subfigure[]{\includegraphics[width=.45\columnwidth]{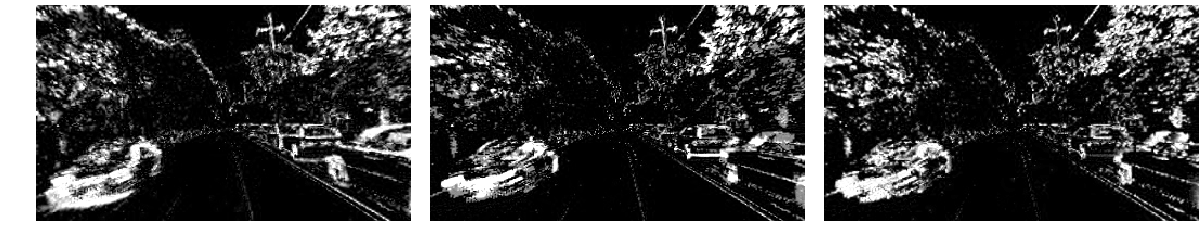}}
%\subfloat(b)
\subfigure[]{\includegraphics[width=.85\columnwidth]{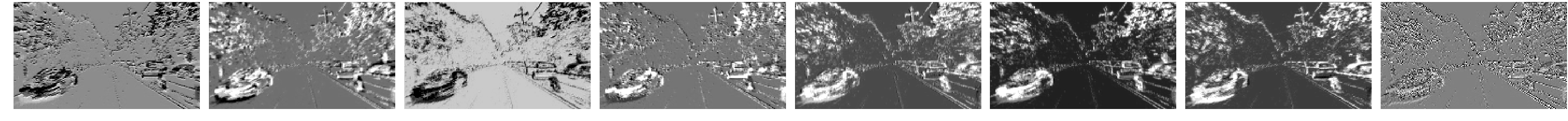}}
%\subfloat(c)
\subfigure[]{\includegraphics[width=.5\columnwidth]{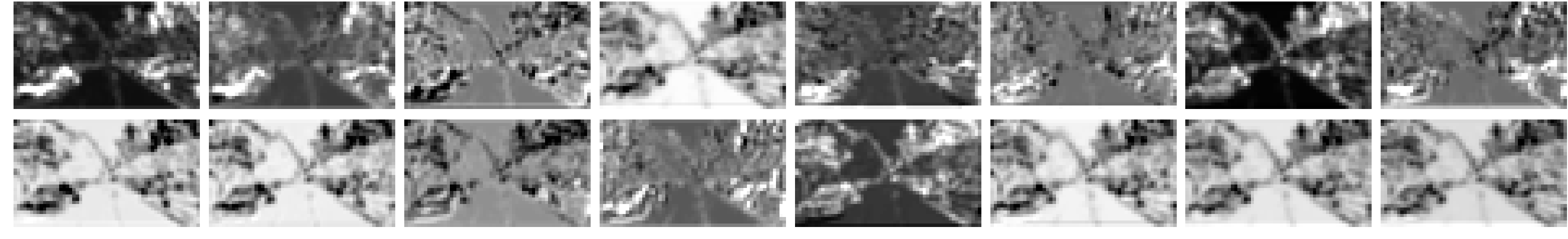}}\\
%\subfloat(d)
\subfigure[]{\includegraphics[width=.25\columnwidth]{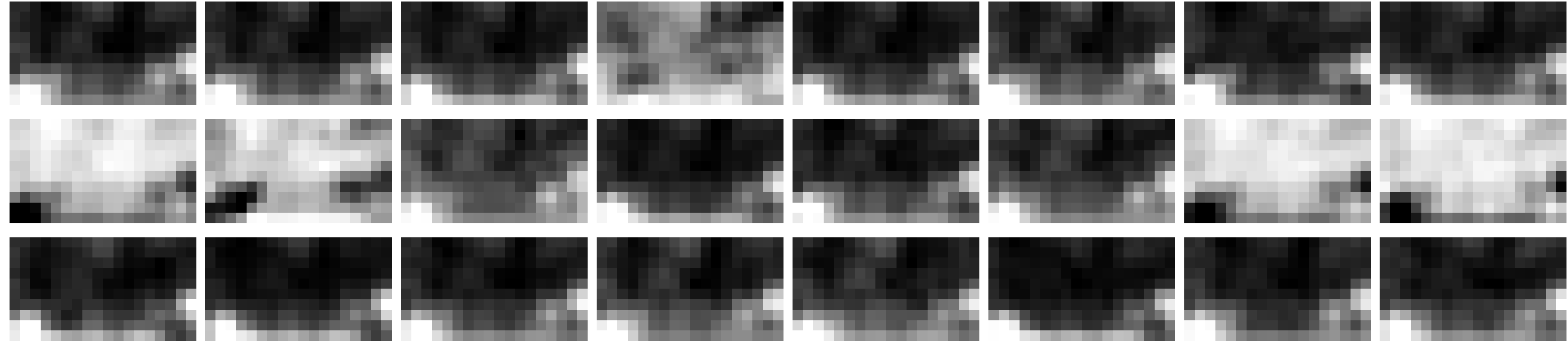}}\\
%\subfloat(e)
\subfigure[]{\includegraphics[width=.5\columnwidth]{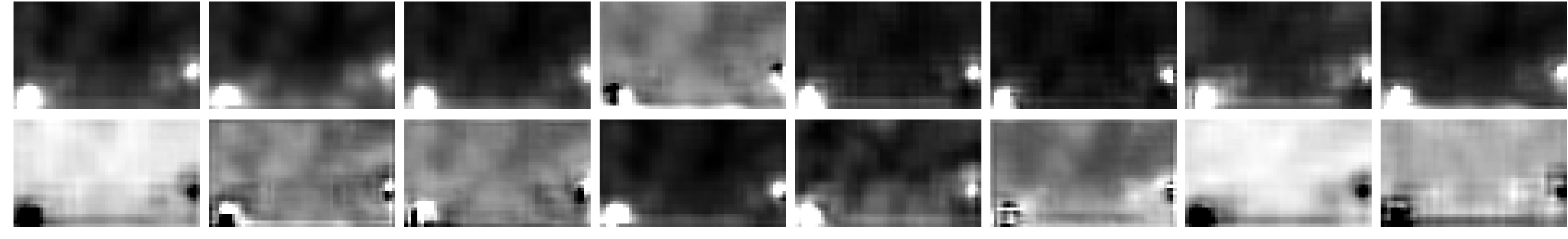}}
%\subfloat(f)
\subfigure[]{\includegraphics[width=.85\columnwidth]{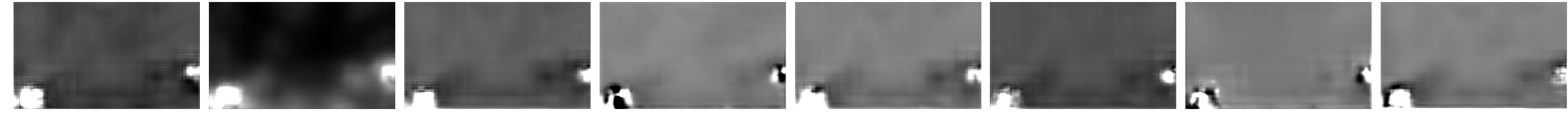}}
%\subfloat(g)
\subfigure[]{\includegraphics[width=.45\columnwidth]{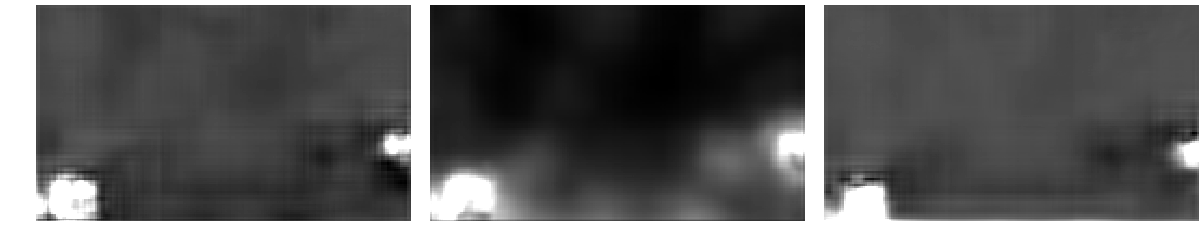}}\\
%\subfloat(h)
\subfigure[]{\includegraphics[width=.08\columnwidth]{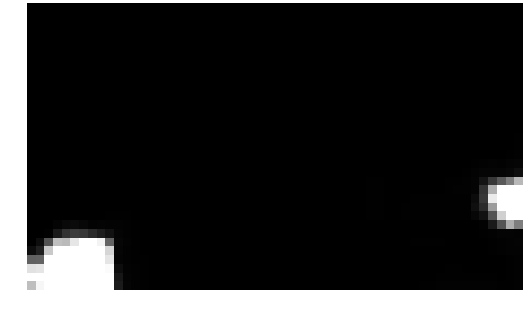}}
%\subfloat(i)
\subfigure[]{\includegraphics[width=.11\columnwidth]{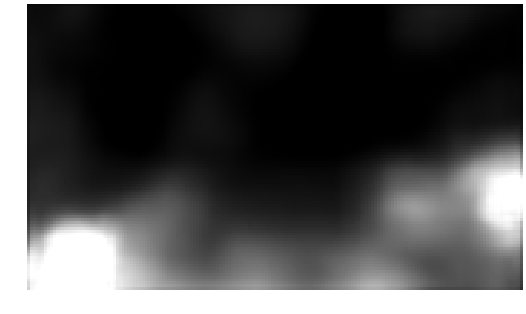}}
%\subfloat(j)
\subfigure[]{\includegraphics[width=.16\columnwidth]{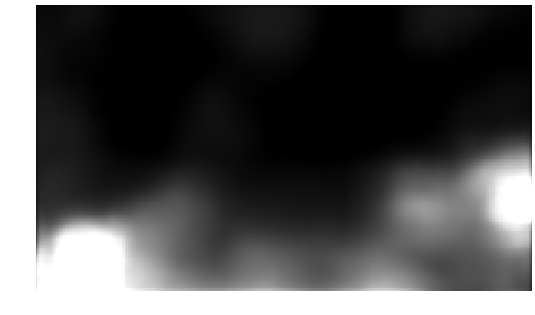}}
\caption{Visualization of feature maps of the depth network. (a) Input channels. (b-g) Feature maps of the encoder-decoder network. (h-j) Multiscale predictions by layers (e-g).}
\label{fig:depthnet_maps}
\end{figure}

\begin{figure}[h]
\centering
%\subfloat(a)
\subfigure[]{\includegraphics[width=.95\columnwidth]{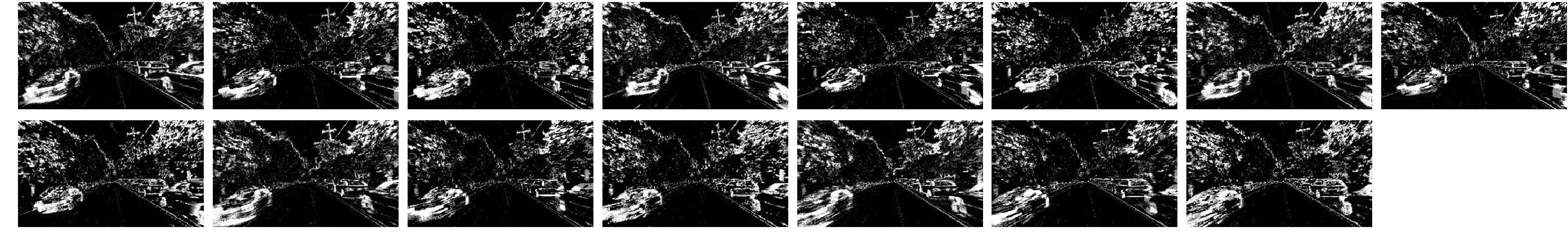}}
%\subfloat(b)
\subfigure[]{\includegraphics[width=.495\columnwidth]{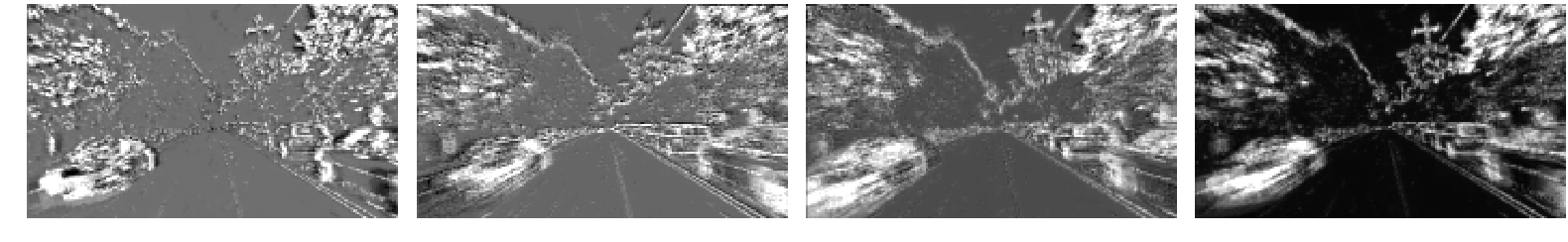}}
%\subfloat(c)
\subfigure[]{\includegraphics[width=.22\columnwidth]{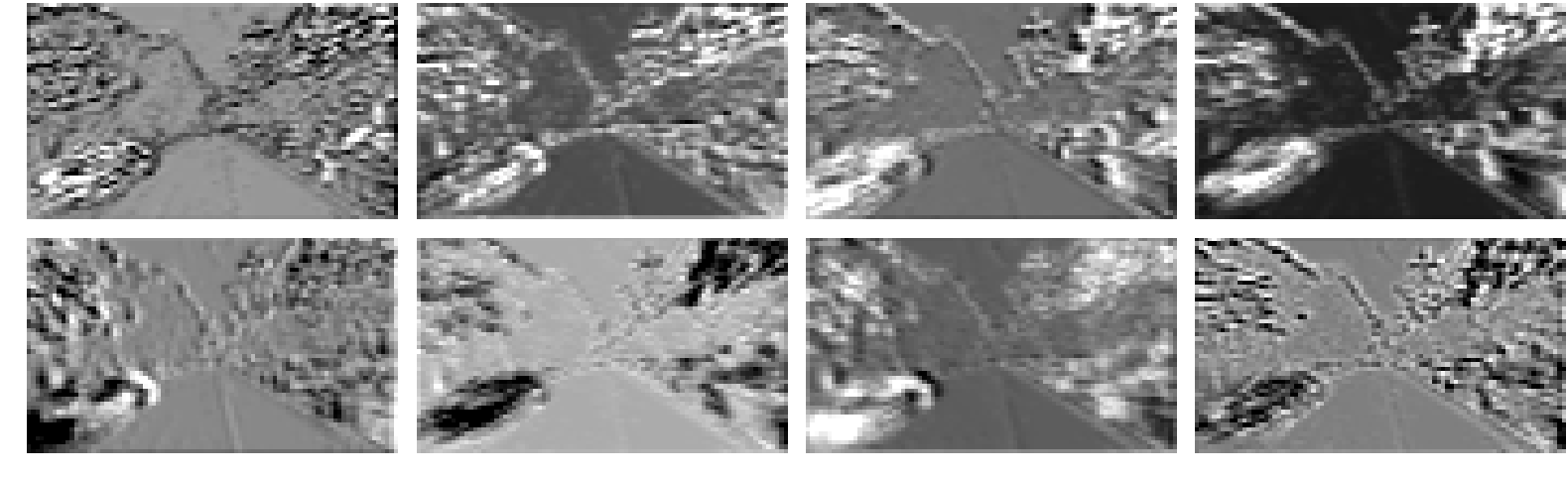}}
%\subfloat(d)
\subfigure[]{\includegraphics[width=.11\columnwidth]{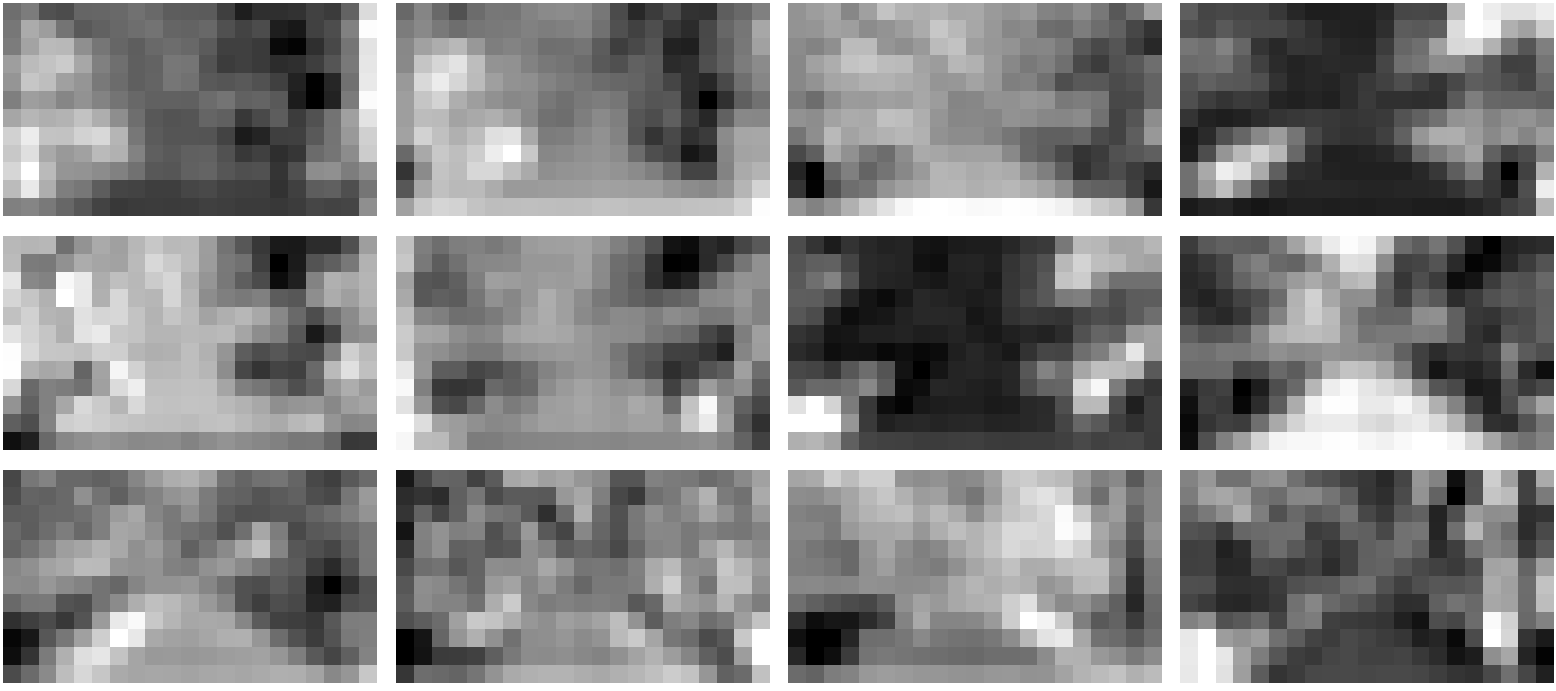}}
\caption{Visualization of the pose network. (a) Input channels. (b-d) Feature maps of the pose network.}
\label{fig:posenet_maps}
\end{figure}

Our lightweight multi-level, multi-resolution design allows us to construct networks of any depth and size. As a preliminary attempt, we set the scaling rate to $1/3$ and $3.0$ for encoding and decoding layers respectively, so the network has only 3 encoding/decoding layers. As the network is small, we can directly visualize all its internal feature maps. A deeper and wider network would produce a higher quality output but also more feature maps, which we do not list here. In Fig.~\ref{fig:depthnet_maps} we have listed all the feature maps in the small depth network. The row number corresponds to the level number of the features for each figure. We notice the encoder seems to play a feature extraction role in the network and the decoder starts to produce semantically meaningful representation corresponding to the desired output (depth). By scrutinizing the pose network outputs, we notice the network is intelligent enough to aggregate information corresponding to different time periods of the events in the first layer (Fig.~\ref{fig:posenet_maps}). Otherwise, mixing up the events at different time period would make the pose estimation harder.

\subsubsection{Performance Versus Event Rate}

\begin{figure}[t]
\begin{center}
\includegraphics[width=1.0\columnwidth]{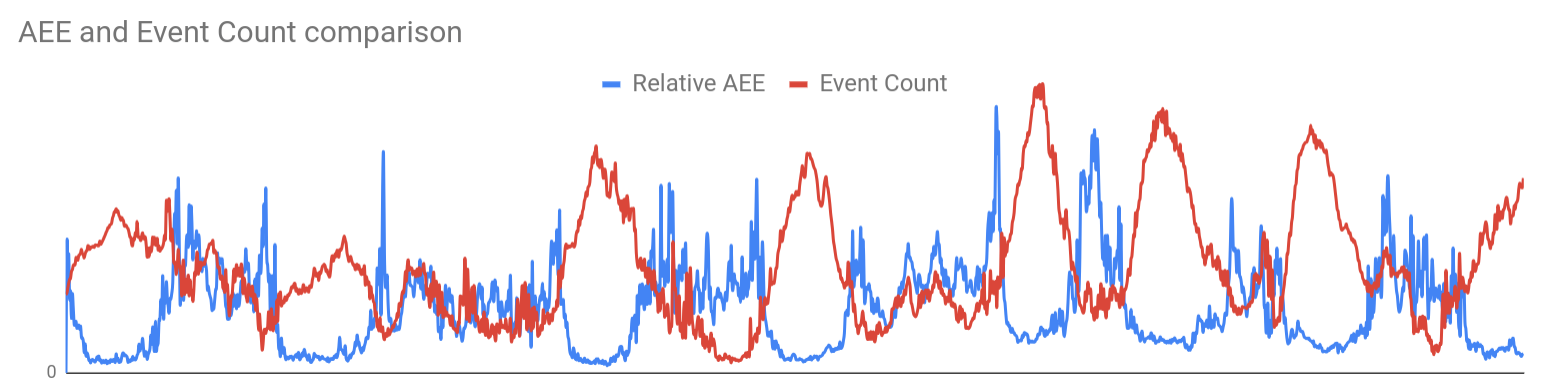}
\end{center}
\vspace{-1.0\baselineskip}
   \caption{\textit{\small{The Average Endpoint Error (blue) and the number of pixels with at least one event (red) for the first 1500 frames of `outdoor_day1' sequence of the MVSEC \cite{MVSEC} dataset. Both plots are normalized so that the mean value is 0.5 for easier comparison.}}}
   \vspace{-0.5\baselineskip}
\label{fig:plot_aee_ecount}
\end{figure}

Since the event data is inherently sparse, we investigate the performance of the pipeline in relation to the data sparsity. 

We measure the data sparsity as a percentage of the pixels on the input images with at least one event. Fig.~\ref{fig:plot_aee_ecount} demonstrates how the event rate is inversely proportional to the average endpoint error for the optical flow (we have observed similar behavior for depth and egomotion). The \textit{outdoor day 1} sequence is used to minimize the influence of the noise.

We find the inverse correlation between event rate and inference quality to be a useful observation, as this property could be efficiently used in sensor fusion in a robotic system. Motivated by that, we provide an additional row to the Table \ref{table:main_table}: $ECN_{erate}$, and report our error metrics once again only for the frames with \textit{higher than average} number of event pixels across the datasets.

\subsection{Qualitative Results}
In addition to the quantitative evaluation, we present a number of samples for qualitative analysis in Fig. \ref{fig:big_pic}. The last three rows of the table show the night sequences, and how the pipeline performs well even when only a few events are available. The third and the fourth rows show indoor scenes. The indoor sequences were relatively small and it is highly possible that the quality of the output would increase given a larger dataset.

\subsection{Optical Flow Evaluation}

We evaluate our optical flow results in terms of Average Endpoint Error ($AEE = \frac{1}{n}\sum \norm{\vec{y} - \vec{y}^*}_2$ with $y^*$ and $y$ the estimated and ground truth value, and $n$ the number of events) and compare our results against two state-of-the-art optical flow methods for event-based cameras: $EV-FlowNet$ \cite{zhu2018ev} and a recent stereo method \cite{UpennSlices} (in the tables - $Zhu18$).

Because our network produces flow and depth values for every image pixel, our evaluation is not constrained by pixels which did not trigger a DVS event. Still, for consistency reasons, we report both numbers for each of our experiments (for example, $ECN$ and $ECN_{masked}$, where the latter has errors computed only on the pixels with at least one event). Similar to KITTI and EV-FlowNet, we report the percentage of outliers - values with error more than 3 pixels or 5\% of the flow vector magnitude.

To compare against \cite{zhu2018ev} and \cite{UpennSlices}, we account for the difference in the frame rates (for example, EV-FlowNet uses the frame rate of the DAVIS classical frames) by scaling our optical flow. We also provide aggregated results for the indoor scenes (split on a train and a test set 80/20 as described above), although these are not the main focus of our study. Our main results are presented in the Table \ref{table:main_table}.

We show that our optical flow performs well during both day and night, all on unseen sequences. The results are typically better for the experiments with event masks except for the \textit{outdoor night}. One possible explanation for that is that this sequence is much noisier with events being generated not only on the edges, which leads to suboptimal masking.

\subsection{Depth Evaluation}
Since there are currently no monocular event-based methods for the depth estimation based on unsupervised learning, we provide the classical scale-invariant depth metrics, used in many works such as \cite{DEPTH_COOLGUY}, \cite{zhou2017unsupervised}, \cite{MetricUser}:
\begin{equation}
Accuracy:  \% of  y_i\ s.t.\ max(\frac{y_i}{y_i^*}, \frac{y_i^*}{y_i}) = \delta < th
\label{eq:delta}
\vspace{-0.5\baselineskip}
\end{equation}
\begin{equation}
	SILog: \frac{1}{n}\sum d_i^2 - \frac{1}{n^2}(\sum d_i)^2, d_i = \log y_i - \log y_i^*
\label{eq:SILog}
\vspace{-0.5\baselineskip}
\end{equation}
\begin{equation}
Absolute Relative Difference: \frac{1}{n}\sum \frac{|y - y^*|}{y^*}
\label{eq:AbsRelDiff}
\vspace{-0.5\baselineskip}
\end{equation}
\begin{equation}
Logarithmic RMSE: \sqrt{\frac{1}{n}\sum \norm{\log y - \log y^*}^2}
\label{eq:RMSElog}
\end{equation}

Our results are presented in Table \ref{table:main_table_depth} for both event count-masked depth values and full, dense depth. Since the night driving scenes have similar depth geometries, we aggregate all 3 sequences in one table entry.

Applying an event mask during the evaluation increases accuracy for all scenes - this is expected, as the inference is indeed more accurate on the pixels with event data. On the contrary, the error rate increases on the outdoor scenes and decreases on the indoor scenes. This is probably due to higher variation of the outdoor scenes and also faster motion of the car.

%\renewcommand{\arraystretch}{1.1}
%\begin{table*}[h]
%\caption{\small{Evaluation of the depth estimation pipeline (results on masked  depth are in braces)}}
%\vspace{-1.0\baselineskip}
%\begin{center}
%\resizebox{1.0\textwidth}{!}{\begin{tabular}{@{\extracolsep{0pt}}lcccccc}
%\hline

%& \multicolumn{1}{l}{} & \multicolumn{2}{l}{Error metric} & \multicolumn{3}{l}{Accuracy metric} \\
% \cline{2-4}\cline{5-7} \rule{0pt}{3ex}
% & Abs Rel & RMSE log & SILog & $\delta < 1.25$  & $\delta < 1.25^2$ & $\delta < 1.25^3$  \\

%\hline\hline
%\textit{outdoor day 1} & 0.29 (0.33) & 0.29 (0.33) & 0.12 (0.14) & 0.80 (0.97) & 0.91 (0.98) & 0.96 (0.99) \\
%\textit{outdoor night 1,2,3}  & 0.34 (0.39) & 0.38 (0.42) & 0.15 (0.18) & 0.67 (0.95) & 0.85 (0.98) & 0.93 (0.99) \\
%\textit{indoor flying} & 0.28 (0.22) & 0.29 (0.25) & 0.11 (0.11) & 0.75 (0.98) & 0.91 (0.99) & 0.96 (1.0) \\ 

%\hline
%\end{tabular}}
%\end{center}
%\vspace{-1.0\baselineskip}

%\label{table:main_table_depth}
%\end{table*}
%\renewcommand{\arraystretch}{1}

\renewcommand{\arraystretch}{1.3}
\begin{table}[]
\caption{\small{Evaluation of the depth estimation pipeline (results on masked, sparce depth are in braces)}}
%\vspace{-1.0\baselineskip}
\begin{tabular}{cllll}
 \hline
\multicolumn{1}{l}{}                      &                               & outdoor day 1 & outdoor night & indoor flying \\ \hline  \hline
\multicolumn{1}{c|}{\multirow{3}{*}{\rotatebox{90}{Error}}} & \multicolumn{1}{l|}{Abs Rel}  & 0.29 (0.33)   & 0.34 (0.39)         & 0.28 (0.22)   \\
\multicolumn{1}{c|}{}                     & \multicolumn{1}{l|}{RMSE log} & 0.29 (0.33)   & 0.38 (0.42)         & 0.29 (0.25)   \\
\multicolumn{1}{c|}{}                     & \multicolumn{1}{l|}{SILog}    & 0.12 (0.14)   & 0.15 (0.18)         & 0.11 (0.11)   \\
\multicolumn{1}{c|}{\multirow{3}{*}{\rotatebox{90}{Accuracy}}} & \multicolumn{1}{l|}{$\delta < 1.25$}       & 0.80 (0.97)   & 0.67 (0.95)         & 0.75 (0.98)   \\
\multicolumn{1}{c|}{}                     & \multicolumn{1}{l|}{$\delta < 1.25^2$}       & 0.91 (0.98)   & 0.85 (0.98)         & 0.91 (0.99)   \\
\multicolumn{1}{c|}{}                     & \multicolumn{1}{l|}{$\delta < 1.25^3$}       & 0.96 (0.99)   & 0.93 (0.99)         & 0.96 (1.0)   
\end{tabular}
\label{table:main_table_depth}
\end{table}
\renewcommand{\arraystretch}{1}

\subsection{Egomotion Estimation}
\label{sec:egomotion}

Our pipeline is capable of inferring egomotion on both day and night sequences, and transfers well from \textit{outdoor day 2} onto \textit{outdoor day 1} and \textit{outdoor night 1,2,3}. Since our pipeline is monocular, we predict the translational component of the velocity up to a scaling factor, while the rotational velocity does not need scaling. Despite our network outputs full 6 degree of freedom velocity, we did not achieve high quality on indoor sequences. This is likely due to highly more complicated motion types and a small size of the indoor dataset. We further discuss this in sec. \ref{sec:discussion}.

\begin{figure}
\centering
\subfigure[\textit{outdoor day 1} sequence]{\includegraphics[width=.95\columnwidth]{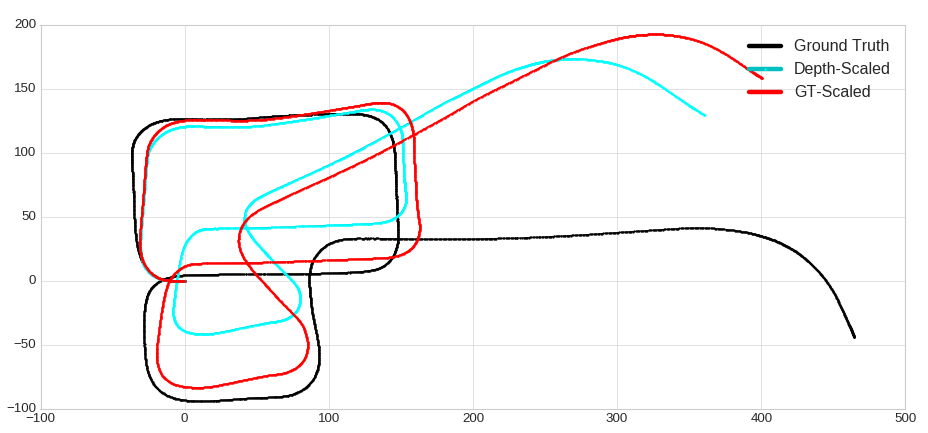}} \\
\subfigure[\textit{outdoor night 2} sequence]{\includegraphics[width=.95\columnwidth]{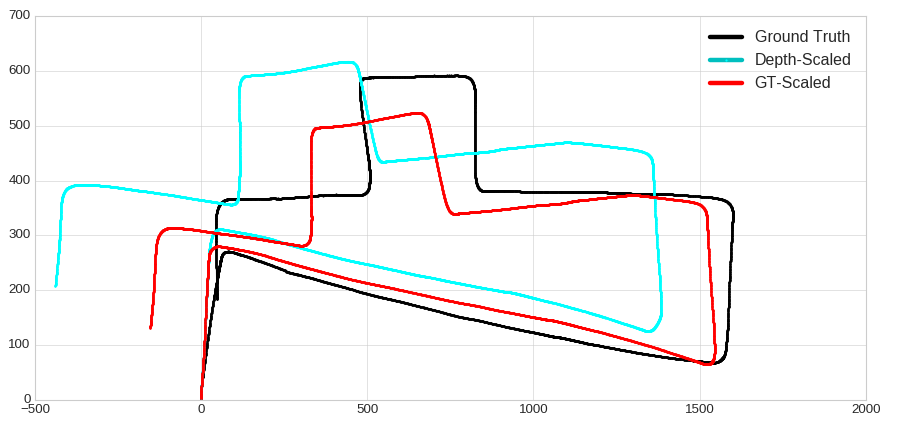}}
\caption{\textit{\small{Estimated trajectories on `outdoor day 1' and `outdoor night 2' sequences, acquired by integrating the egomotion predictions. The network was trained only on `outdoor day 2'. Black: ground truth. Red: network prediction with translational scale applied from ground truth. Cyan: network prediction by assuming the mean depth is fixed throughout the sequence (see sec. \ref{sec:egomotion}) and by applying a single global scaling to the translational pose.}}}
%\vspace{-1.0\baselineskip}
\label{fig:trajectories}
\end{figure}

For the driving scenarios we can \textit{make an important observation} - the mean distance of the road in respect to the camera is often a constant. We crop the lower middle part of the inferred depth image and apply a scaling factor such that the mean depth value (corresponding to the road location) is constant. Consequently, only a single extrinsic value (camera height on the car) is needed to reconstruct the scaled motion. In our experiments, we report egomotion with translational scales taken both from ground truth ($AEE_{tr}^{gt}$) and using the depth constancy constraint ($AEE_{tr}^{depth}$), with a global scale taken from ground truth. The qualitative results can be seen in Fig. \ref{fig:trajectories}.

Unlike \cite{UpennSlices}, we train \textit{SfMlearner} on the event images, and not on the classical frames to allow for evaluation on the night sequences. We provide comparison to the work in \cite{UpennSlices}, although it uses a stereo pipeline and reports results only on the \textit{outdoor day 1} sequence.

To be consistent with \cite{UpennSlices}, we report our trajectory estimation relative pose and relative rotation errors as $RPE = arccos(\frac{t_{pred} \cdot t_{gt}}{\|t_{pred}\|_2 \cdot \|t_{gt}\|_2})$ and $RRE = \|logm(R_{pred}^TR_{gt})\|_2$. Here $logm$ is matrix logarithm and $R$ are Euler rotation matrices. The $RPE$ essentially amounts to the angular error between translational vectors (ignoring the scale), and $RRE$ amounts to the total 3-dimensional angular rotation error. To account for translational scale errors, we report classical Endpoint Errors - computed as a magnitude of the differences in translational component of the velocities. Our quantitative results are presented in Table \ref{table:main_table_egomotion}.

\renewcommand{\arraystretch}{1.6}
\begin{table}[h!]
\caption{Egomotion Estimation Results}
\label{table:main_table_egomotion}
\begin{tabular}{cccccc}
\hline
 &  & \multicolumn{1}{c}{$ARPE$} & \multicolumn{1}{c}{$ARRE$} & \multicolumn{1}{c}{$AEE_{tr}^{gt}$} & \multicolumn{1}{c}{$AEE_{tr}^{depth}$} \\ \hline
\hline

\multicolumn{1}{l|}{$ECN$} & \multicolumn{1}{l|}{\multirow{3}{*}{\rotatebox{90}{outdoor} \rotatebox{90}{   day 1}}} &                      3.98          &                0.00267                &                  0.70                 &                  1.29                    \\
\multicolumn{1}{l|}{$Zhu18$ \cite{UpennSlices}}       & \multicolumn{1}{l|}{}                               & 7.74                           & 0.00867                        & -                                 & -                                    \\
\multicolumn{1}{l|}{$SfMlearner$} & \multicolumn{1}{l|}{}                               & 16.99                          & 0.00916                        & 1.59                                  & 2.39                                     \\ 
\hline

\multicolumn{6}{l}{outdoor night \rule{0pt}{4ex} } \\ 

\hline\hline

\multicolumn{1}{l|}{$ECN$}        & \multicolumn{1}{c|}{\multirow{2}{*}{1}}   & 3.90         &  0.00139         & 0.42             & 1.26                \\
\multicolumn{1}{l|}{$SfMlearner$} & \multicolumn{1}{c|}{}                     & 9.95          & 0.00433          & 1.04             & 2.47               \\

\multicolumn{1}{l|}{$ECN$}        & \multicolumn{1}{c|}{\multirow{2}{*}{2}}   & 3.44          & 0.00202          & 0.80             & 1.34                \\
\multicolumn{1}{l|}{$SfMlearner$} & \multicolumn{1}{c|}{}                     & 13.51          & 0.00499          & 1.66             & 3.15               \\

\multicolumn{1}{l|}{$ECN$}        & \multicolumn{1}{c|}{\multirow{2}{*}{3}}   & 3.28          & 0.00202         & 0.49             & 1.03                \\
\multicolumn{1}{l|}{$SfMlearner$} & \multicolumn{1}{c|}{}                     & 1.053         & 0.00482          & 1.42             & 2.74               

\end{tabular}

\caption*{\textit{\small{Table 3: Egomotion estimation results on driving sequences - `outdoor day 1' and `outdoor night 1,2,3'. The ARPE and ARRE are reported in degrees and radians respectively \cite{UpennSlices}, AEE is in m/s. $AEE_{tr}^{gt}$ - translational endpoint error with ground truth normalization. $AEE_{tr}^{depth}$ - normalized using depth prediction and a global scaling factor (see sec. \ref{sec:egomotion}).}}}
\end{table}
\renewcommand{\arraystretch}{1.0}

\subsection{Discussion and Failure Cases}
\label{sec:discussion}

A monocular pipeline tends infer more information from the shape of the contours on depth estimation and hence would transfer poorly on completely different scenarios. Nevertheless, we were able to achieve good generalization on night sequences and demonstrate promising results for depth and flow for indoor scenes (trained separately on parts of indoor sequences).

\begin{figure}[h]
\begin{center}
\includegraphics[width=1.0\columnwidth]{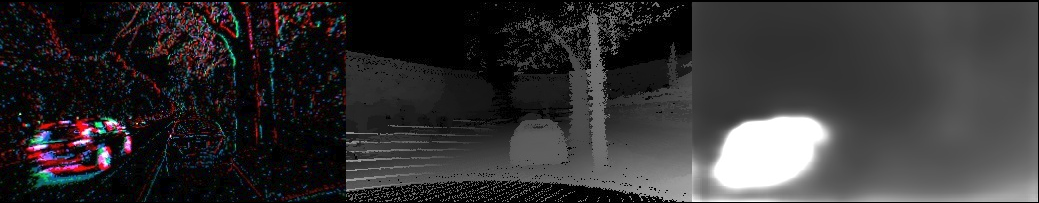}
\end{center}
\vspace{-1.0\baselineskip}
   \caption{\textit{\small{A typical failure case and a dataset artifact: A non-moving car (visible in the middle ground truth inverse depth image) is not visible on the DAVIS camera (left image) which prevents ECN from inferring optical flow or depth correctly (right image is the inference inverse depth image). On the contrary, the moving car on the left side of the road is clearly visible in the event space and its depth inference is correct, but due to the Lidar limitations the depth ground truth is completely missing. This frame is taken from the `outdoor\_night 1' MVSEC sequence.}}}
   \vspace{-0.5\baselineskip}
\label{fig:failure_1}
\end{figure}

We observe a relatively small angular drift on trajectory estimation (Fig. \ref{fig:trajectories}). Despite our model predicting a full 6 degree of freedom
motion we admit that in the car scenario only 2 motion parameters play meaningful role and the network may tend to overfit. For this reason, training on the indoor scenes, featuring more complicated motion would require a notably larger dataset than presented in MVSEC. We still achieve results superior to \textit{SfMlearner} and the stereo method \cite{UpennSlices}, while for the comparison with the latter we must attribute some of our success to the fact that our translational velocity prediction is only up to scale. 
A common shortcoming of event-based sensors in the lack of data when the relative motion is not present. Fig. \ref{fig:failure_1} shows such an example. This issue (although it does not affect egomotion) can be solved only by fusing data from other visual sensors or by moving the event-based sensor continuously. Because of the smoothness constraint used to combat data sparsity, the network tends to blur object boundaries. Still, for the driving environment the contours of obstacles, people and cars are clearly visible, as can be seen in Fig. \ref{fig:big_pic}.

% (see indoor scenario on Fig. \ref{fig:failure_2})

\section{CONCLUSION}

We have presented a novel low-parameter pipeline for generating dense optical flow, depth and egomotion from sparse event camera data. 
%To the best of our knowledge, this is the first deep learning framework to predict depth and optical flow with only events. 
We also have shown experimentally that our new neural network architecture using multi-level features improves upon existing work. Future work will investigate the estimation of moving objects as part of the pipeline, using event cloud representations instead of accumulated events, and the use of space-time frequency representations in the learning.

\bibliography{references.bib}

\begin{thebibliography}{10}

\bibitem{barranco2014contour}
Francisco Barranco, Cornelia Ferm{\"u}ller, and Yiannis Aloimonos.
\newblock Contour motion estimation for asynchronous event-driven cameras.
\newblock {\em Proceedings of the IEEE}, 102(10):1537--1556, 2014.

\bibitem{benosman_event_2014}
R.~Benosman, C.~Clercq, X.~Lagorce, Sio-Hoi Ieng, and C.~Bartolozzi.
\newblock Event-based visual flow.
\newblock {\em Neural Networks and Learning Systems, IEEE Transactions on},
  25(2):407--417, 2014.

\bibitem{benosman_asynchronous_2012}
Ryad Benosman, Sio-Hoi Ieng, Charles Clercq, Chiara Bartolozzi, and Mandyam
  Srinivasan.
\newblock Asynchronous frameless event-based optical flow.
\newblock {\em Neural Netw.}, 27:32 -- 37, March 2012.

\bibitem{delbruck_frame_2008}
T.~Delbruck.
\newblock Frame-free dynamic digital vision.
\newblock In {\em Proceedings of Intl. Symposium on Secure-Life Electronics,
  Advanced Electronics for Quality Life and Society, Tokyo, Japan,}, pages
  21--26, March 2008.

\bibitem{Denman:1976:MSF:2612816.2613024}
Eugene~D. Denman and Alex~N. Beavers, Jr.
\newblock The matrix sign function and computations in systems.
\newblock {\em Appl. Math. Comput.}, 2(1):63--94, January 1976.

\bibitem{DEPTH_COOLGUY}
David Eigen, Christian Puhrsch, and Rob Fergus.
\newblock Depth map prediction from a single image using a multi-scale deep
  network.
\newblock In {\em NIPS}, 2014.

\bibitem{Gallego2018AUC}
Guillermo Gallego, Henri Rebecq, and Davide Scaramuzza.
\newblock A unifying contrast maximization framework for event cameras, with
  applications to motion, depth, and optical flow estimation.
\newblock {\em IEEE Conf. Computer Vision and Pattern Recognition (CVPR)},
  2018.

\bibitem{MetricUser}
Ravi Garg, Vijay Kumar~B. G, and Ian~D. Reid.
\newblock Unsupervised {CNN} for single view depth estimation: Geometry to the
  rescue.
\newblock {\em CoRR}, abs/1603.04992, 2016.

\bibitem{KITTI1}
A.~Geiger, P.~Lenz, and R.~Urtasun.
\newblock Are we ready for autonomous driving? the kitti vision benchmark
  suite.
\newblock In {\em 2012 IEEE Conference on Computer Vision and Pattern
  Recognition}, pages 3354--3361, June 2012.

\bibitem{DBLP:journals/corr/HeZRS15}
Kaiming He, Xiangyu Zhang, Shaoqing Ren, and Jian Sun.
\newblock Deep residual learning for image recognition.
\newblock {\em CoRR}, abs/1512.03385, 2015.

\bibitem{BMVC2016_63}
Guillermo~Gallego Henri~Rebecq and Davide Scaramuzza.
\newblock Emvs: Event-based multi-view stereo.
\newblock In E.~R.~Hancock R.~C.~Wilson and W.~A.~P. Smith, editors, {\em
  Proceedings of the British Machine Vision Conference (BMVC)}, pages
  63.1--63.11, September 2016.

\bibitem{DBLP:journals/corr/IoffeS15}
Sergey Ioffe and Christian Szegedy.
\newblock Batch normalization: Accelerating deep network training by reducing
  internal covariate shift.
\newblock {\em CoRR}, abs/1502.03167, 2015.

\bibitem{10.1007/978-3-319-46466-4_21}
Hanme Kim, Stefan Leutenegger, and Andrew~J. Davison.
\newblock Real-time 3d reconstruction and 6-dof tracking with an event camera.
\newblock In Bastian Leibe, Jiri Matas, Nicu Sebe, and Max Welling, editors,
  {\em Computer Vision -- ECCV 2016}, pages 349--364, Cham, 2016. Springer
  International Publishing.

\bibitem{lichtsteiner_latency_2008}
P.~Lichtsteiner, C.~Posch, and T.~Delbruck.
\newblock A 128 x 128 at 120db 15 micros latency asynchronous temporal contrast
  vision sensor.
\newblock {\em Solid-State Circuits, IEEE Journal of}, 43(2):566--576, 2008.

\bibitem{DBLP:journals/corr/LinM17}
Tsung{-}Yu Lin and Subhransu Maji.
\newblock Improved bilinear pooling with cnns.
\newblock {\em CoRR}, abs/1707.06772, 2017.

\bibitem{liu2017block}
Min Liu and Tobi Delbruck.
\newblock Block-matching optical flow for dynamic vision sensors: Algorithm and
  fpga implementation.
\newblock In {\em Circuits and Systems (ISCAS), 2017 IEEE International
  Symposium on}, pages 1--4. IEEE, 2017.

\bibitem{DBLP:journals/corr/abs-1802-05522}
Reza Mahjourian, Martin Wicke, and Anelia Angelova.
\newblock Unsupervised learning of depth and ego-motion from monocular video
  using 3d geometric constraints.
\newblock {\em CoRR}, abs/1802.05522, 2018.

\bibitem{iROSBetterFlow}
Anton Mitrokhin, Cornelia Fermuller, Chethan Parameshwara, and Yiannis
  Aloimonos.
\newblock Event-based moving object detection and tracking.
\newblock {\em IEEE/RSJ Int. Conf. Intelligent Robots and Systems (IROS)},
  2018.

\bibitem{6891162}
G.~Orchard and R.~Etienne-Cummings.
\newblock Bioinspired visual motion estimation.
\newblock {\em Proceedings of the IEEE}, 102(10):1520--1536, Oct 2014.

\bibitem{DBLP:conf/miccai/RonnebergerFB15}
Olaf Ronneberger, Philipp Fischer, and Thomas Brox.
\newblock U-net: Convolutional networks for biomedical image segmentation.
\newblock In {\em {MICCAI} {(3)}}, volume 9351 of {\em Lecture Notes in
  Computer Science}, pages 234--241. Springer, 2015.

\bibitem{tschechne2014event}
Stephan Tschechne, Tobias Brosch, Roman Sailer, Nora von Egloffstein, Luma~Issa
  Abdul-Kreem, and Heiko Neumann.
\newblock On event-based motion detection and integration.
\newblock In {\em Proceedings of the 8th International Conference on
  Bioinspired Information and Communications Technologies}, pages 298--305,
  2014.

\bibitem{DBLP:journals/corr/abs-1712-00175}
Chaoyang Wang, Jos{\'{e}}~Miguel Buenaposada, Rui Zhu, and Simon Lucey.
\newblock Learning depth from monocular videos using direct methods.
\newblock {\em CoRR}, abs/1712.00175, 2017.

\bibitem{DBLP:journals/corr/abs-1803-08494}
Yuxin Wu and Kaiming He.
\newblock Group normalization.
\newblock {\em CoRR}, abs/1803.08494, 2018.

\bibitem{DBLP:conf/bigdataconf/YeDMFA18}
Chengxi Ye, Chinmaya Devaraj, Michael Maynord, Cornelia Ferm{\"{u}}ller, and
  Yiannis Aloimonos.
\newblock Evenly cascaded convolutional networks.
\newblock In {\em {IEEE} International Conference on Big Data, Big Data 2018,
  Seattle, WA, USA, December 10-13, 2018}, pages 4640--4647, 2018.

\bibitem{zhou2017unsupervised}
Tinghui Zhou, Matthew Brown, Noah Snavely, and David~G. Lowe.
\newblock Unsupervised learning of depth and ego-motion from video.
\newblock In {\em CVPR}, 2017.

\bibitem{zhou2018semi}
Yi~Zhou, Guillermo Gallego, Henri Rebecq, Laurent Kneip, Hongdong Li, and
  Davide Scaramuzza.
\newblock Semi-dense 3d reconstruction with a stereo event camera.
\newblock {\em European Conference on Computer Vision(ECCV)}, 2018.

\bibitem{MVSEC}
A.~Z. Zhu, D.~Thakur, T.~{\"O}zaslan, B.~Pfrommer, V.~Kumar, and K.~Daniilidis.
\newblock The multivehicle stereo event camera dataset: An event camera dataset
  for 3d perception.
\newblock {\em IEEE Robotics and Automation Letters}, 3(3):2032--2039, July
  2018.

\bibitem{zhu2018ev}
Alex~Zihao Zhu, Liangzhe Yuan, Kenneth Chaney, and Kostas Daniilidis.
\newblock Ev-flownet: Self-supervised optical flow estimation for event-based
  cameras.
\newblock {\em Robotics: Science and Systems}, 2018.

\bibitem{UpennSlices}
Alex {Zihao Zhu}, Liangzhe {Yuan}, Kenneth {Chaney}, and Kostas {Daniilidis}.
\newblock {Unsupervised Event-based Learning of Optical Flow, Depth, and
  Egomotion}.
\newblock {\em arXiv e-prints}, page arXiv:1812.08156, Dec 2018.

\end{thebibliography}
\bibliographystyle{plain}

\end{document}